\documentclass[10pt,twocolumn,letterpaper]{article}
\usepackage{cvpr}
\usepackage{times}
\usepackage{epsfig}
\usepackage{graphicx}
\usepackage{amsmath}
\usepackage{amssymb}
\usepackage{algorithm,algorithmicx,algpseudocode}
\usepackage[english]{babel}
\usepackage{authblk}
\usepackage{algcompatible}
\usepackage{booktabs}
\usepackage{multirow}
\usepackage{eqparbox,array}

\usepackage[pagebackref=true,breaklinks=true,letterpaper=true,colorlinks,bookmarks=false]{hyperref}

\cvprfinalcopy 

\def\cvprPaperID{4230} 

\usepackage{lipsum}
\newcommand\blfootnote[1]{%
\begingroup
\renewcommand\thefootnote{}\footnote{#1}%
\addtocounter{footnote}{-1}%
\endgroup
}

\usepackage{authblk}
\makeatletter
\renewcommand\AB@affilsepx{, \protect\Affilfont}
\makeatother

\begin{document}

\title{Bidirectional Graph Reasoning Network for Panoptic Segmentation}
\author[1$\dagger$]{Yangxin Wu}
\author[1$\dagger$]{Gengwei Zhang}
\author[1]{Yiming Gao}
\author[1]{Xiajun Deng}
\author[2]{Ke Gong}
\author[1,2$\star$]{Xiaodan Liang}
\author[1,2]{Liang Lin}
\affil[1]{Sun Yat-sen University}
\affil[2]{DarkMatter AI Research
\protect\\
\textit {\small  \{wuyx29, zhanggw8, gaoym9, dengxj9\}@mail2.sysu.edu.cn, kegong936@gmail.com, xdliang328@gmail.com, linliang@ieee.org}}

\maketitle
\ifcvprfinal\pagestyle{empty}\fi
\ifcvprfinal\thispagestyle{empty}\fi

\pdfoutput=1
\begin{abstract}
   Recent researches on panoptic segmentation resort to a single end-to-end network to combine the tasks of instance segmentation and semantic segmentation. 
   However, prior models only unified the two related tasks at the architectural level via a multi-branch scheme or revealed the underlying correlation between them by unidirectional feature fusion, which disregards the explicit semantic and co-occurrence relations among objects and background. 
   Inspired by the fact that context information is critical to recognize and localize the objects, and inclusive object details are significant to parse the background scene, we thus investigate on explicitly modeling the correlations between object and background to achieve a holistic understanding of an image in the panoptic segmentation task. 
We introduce a Bidirectional Graph Reasoning Network (BGRNet), which incorporates graph structure into the conventional panoptic segmentation network to mine the intra-modular and inter-modular relations within and between foreground \textit{things} and background \textit{stuff} classes. 
In particular, BGRNet first constructs image-specific graphs in both instance and semantic segmentation branches that enable flexible reasoning at the proposal level and class level, respectively.
To establish the correlations between separate branches and fully leverage the complementary relations between \textit{things} and \textit{stuff}, we propose a Bidirectional Graph Connection Module to diffuse  information across branches in a learnable fashion. 
Experimental results demonstrate the superiority of our BGRNet that achieves the new state-of-the-art performance on challenging COCO and ADE20K panoptic segmentation benchmarks.
\end{abstract}
\section{Introduction}
\begin{figure*}[ht]
    \centering
    \includegraphics[width=0.9\linewidth]{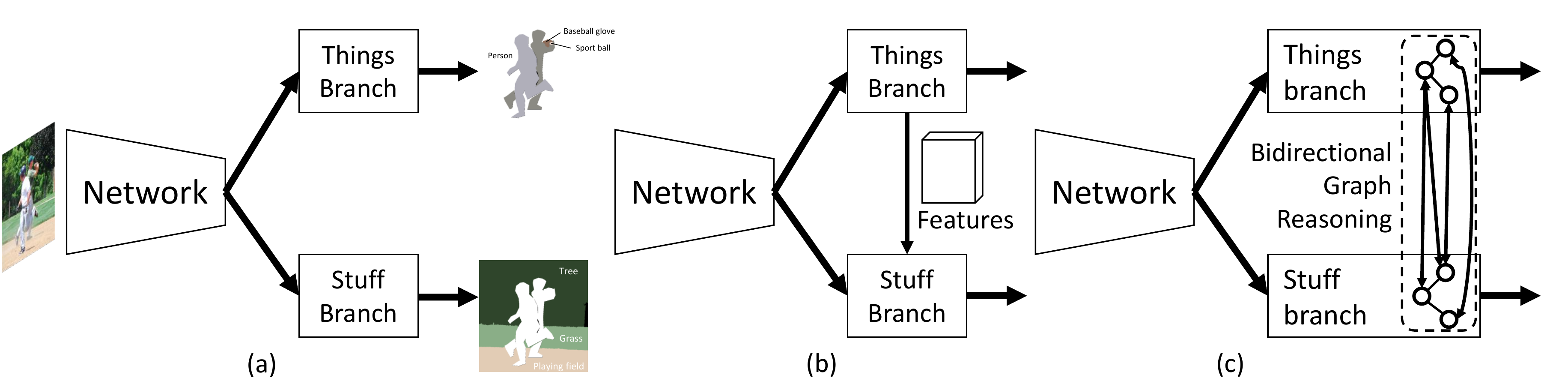}
    \caption{Different architectures for panoptic segmentation. (a) Simple multi-branch structure~\cite{kirillov2019panopticFPN,Liu2019oanet} where two branches have no connection. (b) Unidirectional feature connection structure~\cite{Li_2019_Aunet} that propagates information from \textit{things} branch to \textit{stuff} branch. (c) Our Bidirectional Graph Reasoning Network that enables mutual interaction and promotion for \textit{things} and \textit{stuff} based on graph convolution.}
    \label{fig:motivation}
\end{figure*}

\blfootnote{$\dagger$ Equal contribution. $\star$ Corresponding Author.}
Thanks to the visual reasoning based on  human commonsense, humans are capable of accomplishing recognition and segmentation of the objects and background of an image at a single glance. 
Recent researches have been devoted to developing numerous specific models for instance segmentation~\cite{dai2016instanceFCN,li2017FC_instances} and semantic segmentation~\cite{long2015fcn}. Generally, instance segmentation detects and segments each foreground object (named \textit{things}) while semantic segmentation parses amorphous regions and background (named \textit{stuff}). Tackling the two correlated tasks in separate models, these methods have sacrificed the holistic understanding of an image. 
Recently, a new proposed panoptic segmentation task has attracted researches~\cite{kirillov2019panopticFPN,kirillov2018panoptic,Li_2019_Aunet,Liu2019oanet} to develop end-to-end networks to segment all foreground objects and background contents at the same time. As shown in Figure~\ref{fig:motivation}(a, b), some of the previous works~\cite{kirillov2019panopticFPN,kirillov2018panoptic} unified instance segmentation and semantic segmentation at the architectural level via a multi-branch scheme. The others moved forward to reveal the underlying connection between the two related tasks by unidirectional feature fusion~\cite{Li_2019_Aunet}. 
Although successfully tackling two tasks in one network, these approaches overlooked the explicit semantic and co-occurrence relations between objects and background in a complicated environment, which leads to limited performance gain.

To address these realistic challenges, we reconsider the characteristics of object segmentation as well as scene parsing and investigate on robustly modeling the various relations between them to better tackle the panoptic segmentation task. 
Intuitively, visual context is essential for instance segmentation when predicting fine-grained objects categories and contours~\cite{dvornik2018modeling}, while foreground object details can benefit the segmentation of global scene and stuff~\cite{Li_2019_Aunet}. 
It is obvious and remarkable that \textit{things} and \textit{stuff} can benefit each other by information propagation in one unified network to boost the overall performance of panoptic segmentation. Inspired by this, we introduce a new Bidirectional Graph Reasoning Network (named BGRNet) that incorporates graph structure into the conventional panoptic segmentation network to encode the semantic and co-occurrence relations as well as diffuse information between \textit{things} and \textit{stuff}, as shown in Figure~\ref{fig:motivation}(c).

Specifically, taking advantage of graph convolutional networks~\cite{kipf2016semi}, our BGRNet extracts image-specific graphs from a panoptic segmentation pipeline and learns the diverse relations of \textit{things} and \textit{stuff} utilizing a multi-head attention mechanism.
We propose a Bidirectional Graph Connection Module to bridge \textit{things} graph and \textit{stuff} graph in different branches, which enables graph reasoning and information propagation in a bidirectional way. 
Then we refine the feature representations in both branches by projecting the diffused graph node features.
In this way, BGRNet is aware of the reciprocal relations between \textit{things} and \textit{stuff} and exhibits superior performance in panoptic segmentation.

Furthermore, our BGRNet can be easily instantiated to various network backbones and optimized in an end-to-end fashion. We perform extensive experiments on two challenging panoptic segmentation benchmarks, i.e., COCO~\cite{coco_dataset} and ADE20K~\cite{ade_dataset}. Our approach shows the superior flexibility and effectiveness in modeling and utilizing the relations between \textit{things} and \textit{stuff}, which achieves state-of-the-art performance in terms of PQ on two benchmarks.

\begin{figure*}[ht]
    \centering
    \includegraphics[width=1.0\linewidth]{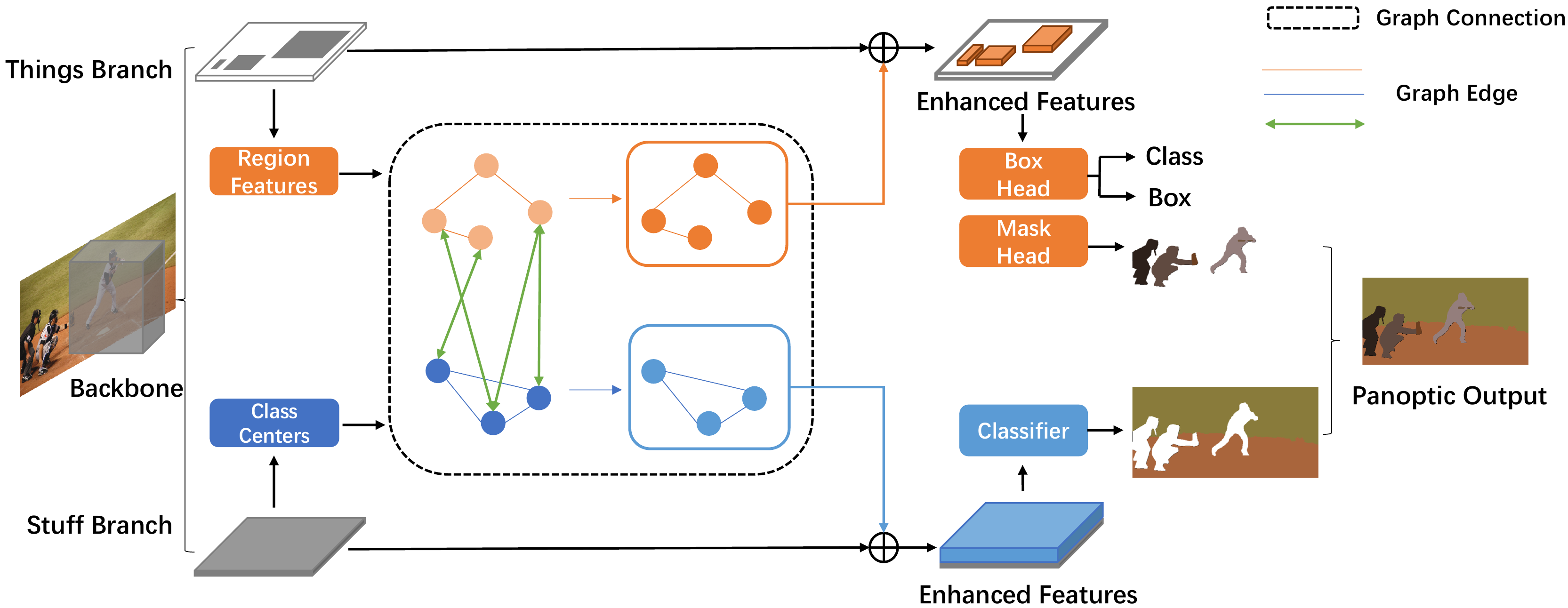}
    \caption{An overview of our BGRNet that can be stacked on any existing two branches panoptic segmentation network. 
    The image features extracted by deep convolutional networks are fed into \textit{things} branch and \textit{stuff} branch. 
    We construct Thing-Graph based on the region features after pooling.
    And we obtain Stuff-Graph node representations by extracting class centers from local feature. 
    Then Bidirectional Graph Connection Module is used to propagate the high-level semantic graph representations within separate branches and across branches.  
    Finally, we re-project the graph features to enhance the discriminability of visual features and improve the performance of both \textit{things} and \textit{stuff} branch.}
    \label{fig:overview}
\end{figure*}

\section{Related Work}

\noindent\textbf{Instance Segmentation.} 
Instance segmentation mainly focuses on locating and segmenting each foreground object. Early methods~\cite{dai2015convolutional_joint_object,hariharan2014simultaneous} followed a bottom-up scheme~\cite{arbelaez2014multiscale} or top-down scheme based on segment proposals~\cite{hariharan2015hypercolumns_ojbect}, until Mask R-CNN~\cite{he2017mask} extended Fast R-CNN to deal with instance segmentation by predicting instance masks and class labels in parallel, which became a common backbone for instance segmentation. Mask Scoring R-CNN ~\cite{huang2019mask} corrected Mask R-CNN by aligning mask quality with mask score.

\noindent\textbf{Semantic Segmentation.} 
Semantic segmentation parses scene images into per-pixel semantic classes. Began with FCNs~\cite{long2015fcn} and DeepLab family~\cite{chen2018encoder}, methods like fully convolutional network and atrous convolution made semantic segmentation thriving by boosting the overall segmentation quality. Besides, the scene parsing method with global context information was also studied in ~\cite{yuan2018ocnet,zhao2017psp}.

\noindent\textbf{Panoptic Segmentation.}
Panoptic Segmentation, a novel task introduced by~\cite{kirillov2018panoptic}, has lately received extensive attention by researchers. 
The task, which unifies instance segmentation and semantic segmentation, requires an algorithm that can segment foreground instances and background semantic classes simultaneously.  In ~\cite{kirillov2018panoptic}, Kirillov \textit{et al}. simply combined the results from PSPNet and Mask R-CNN heuristically to produce panoptic segmentation outputs. 
Not long after, ~\cite{kirillov2019panopticFPN} proposed an end-to-end network for the panoptic task with a shared backbone and two branches: thing branch for instance segmentation and stuff branch for semantic segmentation, respectively.  
Instead of learning two tasks separately, ~\cite{Li_2019_Aunet} tried to utilize the features of the instance segmentation branch to boost the performance of the semantic segmentation branch through an attention mechanism. 
~\cite{Liu2019oanet} proposed a spatial ranking module, to address the occlusion problem which hinders the performance of panoptic segmentation. 
Moreover, UPSNet~\cite{xiong2019upsnet} made use of deformable convolutions together with a parameter-free panoptic head in pursuit of more performance gain. 
A mini-deeplab module was also used to capture more contextual information in ~\cite{porzi2019seamless}.

\noindent\textbf{Graph Reasoning.}
There have been a surge of interest in graph-based methods~\cite{kipf2016semi,velivckovic2017graph,xu2018powerful,yang2019graph,chen2019graphFail} and graph reasoning has shown to have substantial practical merit for many tasks through modeling the domain knowledge in a single graph~\cite{chen2019graph,jiang2018hybrid,wang2018non,gong2019graphonomy} or directly fusing the graph reasoning results~\cite{fu2019dual}. 
However, the mainstream approaches of panoptic segmentation are lack of the investigation on mining mutual relations from different domains (e.g.\, position and channel reasoning in network, \textit{things} and \textit{stuff} subsets) since different graph subsets need more explicit connections for mutual interaction and promotion. 
In this paper, we propose Bidirectional Graph Reasoning that propagates information from different graphs to support more flexible and complex reasoning tasks in general cases. 
Moreover, different from ~\cite{chen2019graph,jiang2018hybrid,wang2018non} that use a single graph for reasoning, our method aims to build a Graph Connection Module, whose nodes have strong semantics (rather than ambiguous nodes in ~\cite{chen2019graph}) and are hence more explainable and capable of encoding various relations.

\section{Bidirectional Graph Reasoning Network}

\subsection{Overview}
\begin{figure*}[ht]
    \centering
    \includegraphics[width=1.0\linewidth]{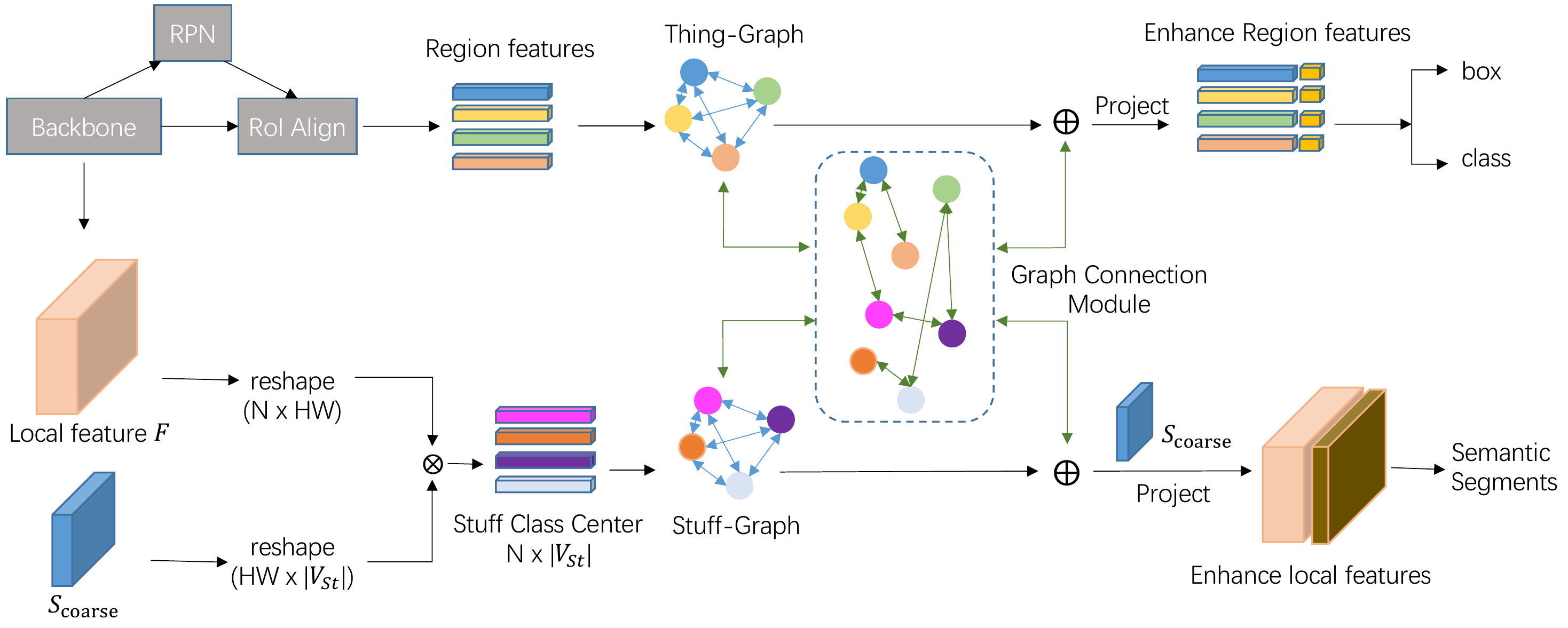}
    \caption{Diagram of our intra-modular graph, i.e., Thing-Graph and Stuff-Graph, and our inter-modular Graph Connection module. 
    For Thing-Graph, we utilize the pooled region features as region graph nodes. 
    For Stuff-Graph, we extract class center from local feature via coarse score map. 
    Then Graph Connection Module diffuses information across various graph nodes and intra-modular graph reasoning is performed to project graph nodes features to visual features at the proposal and pixel level, respectively, in order to refine the results of instance segmentation and semantic segmentation, which are then heuristically combined in an NMS-like procedure following \cite{kirillov2018panoptic}. 
    }
    \label{fig:graph_construct}
\end{figure*}

The panoptic segmentation task is to assign each pixel in an image a semantic label and an instance id.
Current methods typically address this issue with a unified model using two branches for foreground \textit{things} and background \textit{stuff} separately \cite{de2018panoptic_joint,kirillov2019panopticFPN,li2018learning_fuse,Li_2019_Aunet}.
In detail, for an input image, the final panoptic segmentation result was generated by combining results from two branches using fusion strategy following \cite{kirillov2018panoptic}. 
Extending the simple but effective baseline in \cite{kirillov2019panopticFPN}, we aim at further mining the intra-branch and inter-branch relations within and between foreground \textit{things} or background \textit{stuff}.
Firstly, as shown in Figure \ref{fig:graph_construct}, we build image-specific graphs in two separate branches in the network to enable flexible reasoning at the proposal level and class level. 
In the instance segmentation branch, a region graph is established to capture the pair-wise relationships among proposals. 
In the semantic segmentation branch, we build a graph based on the extracted class center that allows efficient global reasoning in a coarse-to-fine paradigm.
Secondly, we propose a Bidirectional Graph Connection Module to deduce the implicit semantic relations between \textit{things} and \textit{stuff} in a learnable fashion. 
After diffusing information across various nodes, intra-modular reasoning is performed to refine the visual features of two branches. 
In this way, we explicitly model the correlations between \textit{things} and \textit{stuff} class and leverage their complementary relations in a global view, 
which facilitates panoptic segmentation and has substantial practical merit in our experiments. 
An overview of our Bidirectional Graph Reasoning Network is shown in Figure~\ref{fig:overview}.

\subsection{Graph Representation}
\label{sec_graph_repre}
Formally, we define a graph as $G = (V,A,X)$ where $V$ is the set of nodes, $A$ denotes the adjacency matrix and 
$X$ is the feature matrix where each row corresponds to a node in $V$.

\textbf{Building Thing-Graph.}
In the classic object detection paradigm, extracted regions are analyzed separately without considering the underlying dependencies between objects, which leads to inconsistent detection results and limited performance in more challenging tasks like panoptic segmentation. 
To remedy this issue, we introduce a Thing-Graph to reason directly beyond local regions, which can refine visual features of certain regions that suffer from occlusions, class ambiguities and tiny-size objects.
Specifically, we build a Thing-Graph $G_{th} = (V_{th}, A_{th}, X_{th})$ on each input image, where $|V_{th}|$ equals to the number of detected regions in the image, $X_{th} \in \mathbb{R}^{|V_{th}| \times N}$ are extracted features from backbone of all regions 
and $N$ is the dimension of the region feature.
Considering the diverse relations among regions, we render the edges in $G_{th}$ learnable to allow flexible reasoning among multiple proposals. 
We also demonstrate the effectiveness of this learnable scheme by comparing the results of using different kinds of knowledge graphs in Section \ref{ablation}.

\textbf{Building Stuff-Graph.} 
As for semantic segmentation, a naive idea of building a Stuff-Graph can be considering each pixel as a graph node similar to the non-local network \cite{WangNon}. 
However, this approach exhibits clear limitations in dense predictions of semantic segmentation since it requires a large amount of computation and vast GPU memory occupation. 
Thus, to reduce the computation overhead as well as capture the long-range dependencies, we project the entire feature map to the vertices of Stuff-Graph so that every vertex represents a specific \textit{stuff} class. 
Regarding Stuff-Graph $G_{st} = (V_{st}, A_{st}, X_{st})$, given the coarse score map $S_{coarse} \in \mathbb{R}^{|V_{st}| \times H \times W}$ produced by the original segmentation head in the baseline network, and segmentation feature map $F \in \mathbb{R}^{N \times H \times W}$, where $N$ is the number of feature channels, we first reshape $S_{coarse}$ to $\mathbb{R}^{HW \times |V_{st}|}$ and $F$ to $\mathbb{R}^{N \times HW}$. 
After performing softmax along the $HW$ channel on score map, we can obtain class nodes feature $X_{st} \in \mathbb{R}^{|V_{st}| \times N}$ by matrix multiplication and transposition:
\begin{equation} \label{eq_cls_center}
    X_{st} = (\bar{F}  \bar{S}_{coarse})^{T},
\end{equation}
where $\bar{F}$ and $\bar{S}_{coarse}$ represent $F$ and  $S_{coarse}$ after reshaping. The intuition behind Equation \ref{eq_cls_center} is that local features, i.e., the features of pixels, are gathered to obtain class nodes feature based on pixel affinity via soft-mapping. 
By assigning global class nodes features to $X_{st}$, we significantly reduce computation overhead in building a Stuff-Graph since $HW \gg |V_{st}|$. 
Besides, the extracted \textit{stuff} nodes are more representative and can provide global clues to further benefit the final classification process after remapping them to local features. 
We further demonstrate the representative characteristics of the extracted class centers in Stuff-Graph in Section \ref{ablation}.
The processes of building Thing-Graph and Stuff-Graph are visualized in Figure \ref{fig:graph_construct}.

\subsection{Bidirectional Graph Connection Module}
\label{graph_conn_sub}
Given the Thing-Graph and Stuff-Graph, we aim to model the mutual relations between \textit{things} and \textit{stuff} and propagate the features across all nodes in both $G_{th}$ and $G_{st}$.
The rationale behind the design of graph nodes feature fusion module across branches is quite straightforward and comprehensible since there exists a consistent pattern of the co-occurrence of foreground \textit{things} and background \textit{stuff} in real-world scenarios.
For example, when there exist objects like \textit{persons}, \textit{sports balls}, \textit{baseball bats} and \textit{baseball gloves} in an image, it is more reasonable to predict the stuff of \textit{sand} and \textit{playing field}, and vice versa.
Therefore, we distill this insight into Graph Connection Module to bridge all semantic information across branches (between foreground \textit{things} and background \textit{stuff}).
In this way, the information, relations or visual correlations of different categories from separate branches can be exploited.

The Graph Connection from Thing-Graph to Stuff-Graph can be formulated as:
\begin{equation}
    X_{t-s} = A_{t-s}X_{th}W_{st},
\end{equation}
where $A_{t-s}\in \mathbb{R}^{|V_{st}|\times |V_{th}|}$ is a transfer matrix for propagating the information from Thing-Graph to Stuff-Graph, $W_{st}\in \mathbb{R}^{N \times D_0}$ is a trainable projection matrix.
$X_{t-s}$ is the mapped node features from Thing-Graph to Stuff-Graph.
Similarly, the Graph Connection from Stuff-Graph to Thing-Graph can be obtained utilizing $X_{st}$ and transfer matrix $A_{s-t}$ with a trainable matrix $W_{th}$.
Therefore, we seek for appropriate transfer matrix $A_{t-s}=\{a_{ij}^{t-s}\}$ and $A_{s-t}=\{a_{ij}^{s-t}\}\in \mathbb{R}^{|V_{th}|\times |V_{st}|}$, where $a_{ij}^{s-t}$ denotes the connection weight from the $j^{th}$ node of Stuff-Graph to the $i^{th}$ node of Thing-Graph.

Based on the graph representation and Graph Connection, our graph structure can be naturally decomposed into blocks, given by
\begin{equation}
\label{eq_block}
    \hat{\mathbf{A}}=\left[\begin{array}{cc}{A_{t h}} & {A_{s-t}} \\ {A_{t-s}} & {A_{s t}}\end{array}\right]
    ,
    \hat{\mathbf{X}}=\left[\begin{array}{l}{X_{t h}} \\ {X_{s t}}\end{array}\right],
\end{equation}
where $A_{th},A_{st},A_{t-s},A_{s-t}$ are normalized adjacency matrices for thing-to-thing pairs, stuff-to-stuff pairs, thing-to-stuff pairs, and stuff-to-thing pairs respectively. 
To model the distribution of different node features and adaptively handle their pairwise relations, we resort to attention mechanism \cite{velivckovic2017graph} to obtain sufficient expressive power in our model.
Formally, for any two nodes $x_i, x_j$ in $\hat{\mathbf{X}}$, the edge weight $\alpha_{ij}$ is computed by:
\begin{equation}
    \alpha_{i j}=\frac{\exp \left( \delta \left(W\left[ x_i \|  x_j\right]\right)\right)}{\sum_{k \in \mathcal{N}_{i}} \exp \left(\delta \left(W\left[ x_i \|  x_k\right]\right)\right)},
\end{equation}
where $||$ is the concatenation operation, $\mathcal{N}_i$ is the neighborhood of node $i$, $\delta$ is LeakyReLU nonlinear activation function, and $W$ is weight matrix. 
For simplicity, we build a fully connected graph for $\hat{\mathbf{X}}$, i.e.,  $\mathcal{N}_i$ contains all nodes in $\hat{\mathbf{X}}$.

\noindent\textbf{Updating node features.}
Formally, with normalized graph adjacency matrix $\hat{\mathbf{A}}$ and node features $\hat{\mathbf{X}}$, a single graph reasoning layer is given by
\begin{equation}
    \widetilde{\mathbf{X}} = \left[\begin{array}{c}{\widetilde{X}_{th}} \\ {\widetilde{X}_{st}}\end{array}\right] = \hat{\mathbf{X}} \oplus \sigma(\hat{\mathbf{A}}\hat{\mathbf{X}}\otimes \hat{\mathbf{W}}),
\end{equation}
where
\begin{equation}
    \hat{\mathbf{W}}=\left[\begin{array}{c}{W_{t h}} \\ {W_{s t}}\end{array}\right],
    \hat{\mathbf{X}} \otimes \hat{\mathbf{W}}=\left[\begin{array}{c}{X_{t h} W_{t h}} \\ {X_{s t} W_{s t}}\end{array}\right], 
\end{equation}
$W_{th},W_{st}\in \mathbb{R}^{D_0\times D_0}$ are trainable weight matrices, $\widetilde{X}_{th}$, $\widetilde{X}_{st}$ are node features of new Thing-Graph and Stuff-Graph respectively, $\oplus$ denotes concatenation, and $\sigma$ is ReLU nonlinear function.
Using $T$ Graph Reasoning layers, the model will propagate and update the information among classes to build more discriminating representations.

\subsection{Project Nodes Features to Visual Features}
\label{sec_proj}
To refine the results of instance and semantic segmentation, we project graph nodes features to visual features at the proposal and pixel level, respectively. 
We illustrate this process in Figure \ref{fig:graph_construct}.

\noindent\textbf{Intra-modular reasoning for detection.}
When enhancing the features of \textit{things} branch, we only care about the features in proposals.
Hence we concatenate the updated Thing-Graph features to each proposal after adjusting their dimension:
\begin{equation}
    f_{th} = A_{th}\tilde{X}_{th}W_{th}^{intra},
\end{equation}
where $W_{th}^{intra} \in \mathbb{R}^{(N+D_0)\times D_1}$ is the weight matrix for intra-modular reasoning in \textit{things} branch.
Then we concatenate enhanced features $f_{th}$ to the visual features of proposals and feed them into the final fully connected layer to obtain the detection results. 

\noindent\textbf{Intra-modular reasoning for segmentation.}
To facilitate the dense prediction in the \textit{stuff} branch, we need to enhance the local feature of each pixel under the guidance of extracted class centers. 
This can be regarded as the inverse operation of Equation \ref{eq_cls_center}.
We reshape $S_{coarse}$ to $\mathbb{R}^{HW \times |V_{st}|}$, the enhanced feature of \textit{stuff} branch can be calculated as:
\begin{equation}
    f_{st} = S_{coarse}\tilde{{X}}_{st}W_{st}^{intra},
\end{equation}
where $W_{st}^{intra} \in \mathbb{R}^{(N+D_0)\times D_2}$ is the weight matrix for intra-modular reasoning in \textit{stuff} branch.
Then $f_{st}$ is concatenated with local feature $F$, which is then fed into the final convolution layer to obtain semantic segmentation results.


\section{Experiments}

\subsection{Experimental Settings}

\noindent\textbf{Implementation Details.} 
The architecture of BGRNet is built on Mask R-CNN~\cite{he2017mask} with a simple semantic segmentation branch similar to \cite{xiong2019upsnet}.
To be exact, the multi-level features from ResNet50-FPN~\cite{He2015resnet,lin2017fpn} first undergo deformable subnets with 3 convolution layers per level and are then bilinearly upsampled to 1/4 of the original scale of the input image. 
Finally, features from different levels are added together and $1 \times 1$ convolution with softmax is applied to predict all \textit{stuff} classes.
We follow all hyper-parameters settings and data augmentation strategies in Panoptic-FPN~\cite{kirillov2019panopticFPN}. 
We implement our model using Pytorch~\cite{paszke2017automatic} and train all models with 8 GPUs with a batch size of 16. The initial learning rate is 0.02 and is divided by 10 two times during fine-tuning. 
For COCO, we train for 12 epochs, i.e., 1x schedule, following~\cite{kirillov2019panopticFPN}. 
For ADE20K, we train for 24 epochs and keep the learning rate schedule in proportion to COCO. 
We adopt an SGD optimizer with a momentum of 0.9 and a weight decay of 5e-4. 
We find it beneficial to extend the attention mechanism to multi-head attention \cite{velivckovic2017graph} and we applied 3 independent output attention heads.
We use two Graph Reasoning layers (\ie $T=2$) and dimension $N=D_0=D_1=D_2=128$.

\noindent\textbf{Datasets and Evaluation Metrics.}
We evaluate our method on COCO~\cite{coco_dataset} and ADE20K~\cite{ade_dataset}. 
COCO is one of the most challenging datasets for panoptic segmentation consisting of 115k images for training, 5k images for validation, and 20k images for \textit{test-dev} with 80 \textit{things} and 53 \textit{stuff} classes. 
\noindent ADE20K is a densely annotated dataset for panoptic segmentation containing 20k images for training, 2k images for validation and 3k images for test, with 100 \textit{things} and 50 \textit{stuff} classes.
Following~\cite{kirillov2018panoptic}, we adopt \textit{panoptic quality (PQ)}, \textit{semantic quality (SQ)}, and \textit{recognition quality (RQ)} for evaluation.

\begin{table}[t]
\caption{Performance comparisons with the state-of-the-art on the COCO val set. $\dagger$ indicates our implementation.
Panoptic-FPN-D is the deformable counterpart of Panoptic-FPN \cite{kirillov2019panopticFPN}. All methods use ResNet50-FPN as the backbone network.
}
\label{tab:compare_sota_coco}
\centering
\begin{tabular}{c|c|c|cc}
    \toprule[1pt]
    Method  & DF Conv. & PQ & PQ$^{Th}$ & PQ$^{St}$ \\
    \midrule[1pt]
    Panoptic-FPN~\cite{kirillov2019panopticFPN}  &  & 39.0 & 45.9 & 28.7 \\
    Panoptic-FPN-D$^{\dagger}$ & \checkmark & 39.9 & 46.9 & 29.3 \\
    AUNet~\cite{Li_2019_Aunet} &  & 39.6 & 49.1 & 25.2 \\
    OANet~\cite{Liu2019oanet}  &  & 39.0 & 48.3 & 26.6 \\
    
    UPSNet-C~\cite{xiong2019upsnet}  & \checkmark & 41.5 & 47.5 & 32.6 \\
    UPSNet-CP~\cite{xiong2019upsnet}  & \checkmark & 41.5 & 47.3 & 32.8 \\
    UPSNet~\cite{xiong2019upsnet} & \checkmark & 42.5 & 48.5 & 33.4 \\
    SpatialFlow~\cite{chen2019spatialflow}  &  & 40.9 & 46.8 & 31.9 \\

    Our BGRNet        & \checkmark &\textbf{43.2}      &\textbf{49.8}           &\textbf{33.4}          \\ 
\bottomrule[1pt]
\end{tabular}
\end{table}

\begin{table}[ht]
\caption{Performance comparisons on ADE20K val set. Panoptic-FPN-D is the deformable counterpart of Panoptic-FPN \cite{kirillov2019panopticFPN}. $\dagger$ indicates our implementation.}
\label{tab:compare_sota_ade}
\centering
\begin{tabular}{l|lll}
\toprule[1pt]
Methods  & PQ & PQ$^{Th}$ & PQ$^{St}$  \\ 
\midrule[1pt]
Panoptic-FPN$^{\dagger}$~\cite{kirillov2019panopticFPN}       &29.3 &32.5    &22.9     \\   
Panoptic-FPN-D$^{\dagger}$~\cite{kirillov2019panopticFPN}  &30.1 &33.1    &24.0     \\   
Our BGRNet  &  \textbf{31.8}  & \textbf{34.1}   &  \textbf{27.3}  \\
\bottomrule[1pt]

\end{tabular}
\end{table}

\subsection{Comparisons with state-of-the-art}
Comparisons with recent state-of-the-art methods on COCO and ADE20K dataset are listed in Table \ref{tab:compare_sota_coco}, \ref{tab:compare_sota_ade}. 
Some previous methods achieve high performance with over 42.5\% PQ, thanks to the specially designed panoptic head~\cite{Liu2019oanet}, multi-scale information~\cite{kirillov2019panopticFPN,Liu2019oanet}, and two sources of attention~\cite{Li_2019_Aunet}.
Unlike previous methods \cite{xiong2019upsnet,Liu2019oanet,Li_2019_Aunet}, our BGRNet does not rely on complicated feature fusion process, i.e., RoI-Upsample \cite{Li_2019_Aunet}, spatial ranking module \cite{Liu2019oanet}, mask pruning process \cite{xiong2019upsnet}.
Instead, we utilize powerful graph models to capture intra-modular and inter-modular dependencies across separate branches. 
Thus, we achieve  consistent accuracy gain over existed methods and set the new state-of-the-art results in terms of $PQ$, $PQ^{Th}$, $PQ^{St}$.
The advanced results demonstrate the superiority of our BGRNet that incorporates the reciprocal information and deduces underlying relations between \textit{things} and \textit{stuff} appeared in the image.

\begin{figure*}[ht]
    \centering
    \includegraphics[width=1.0\linewidth]{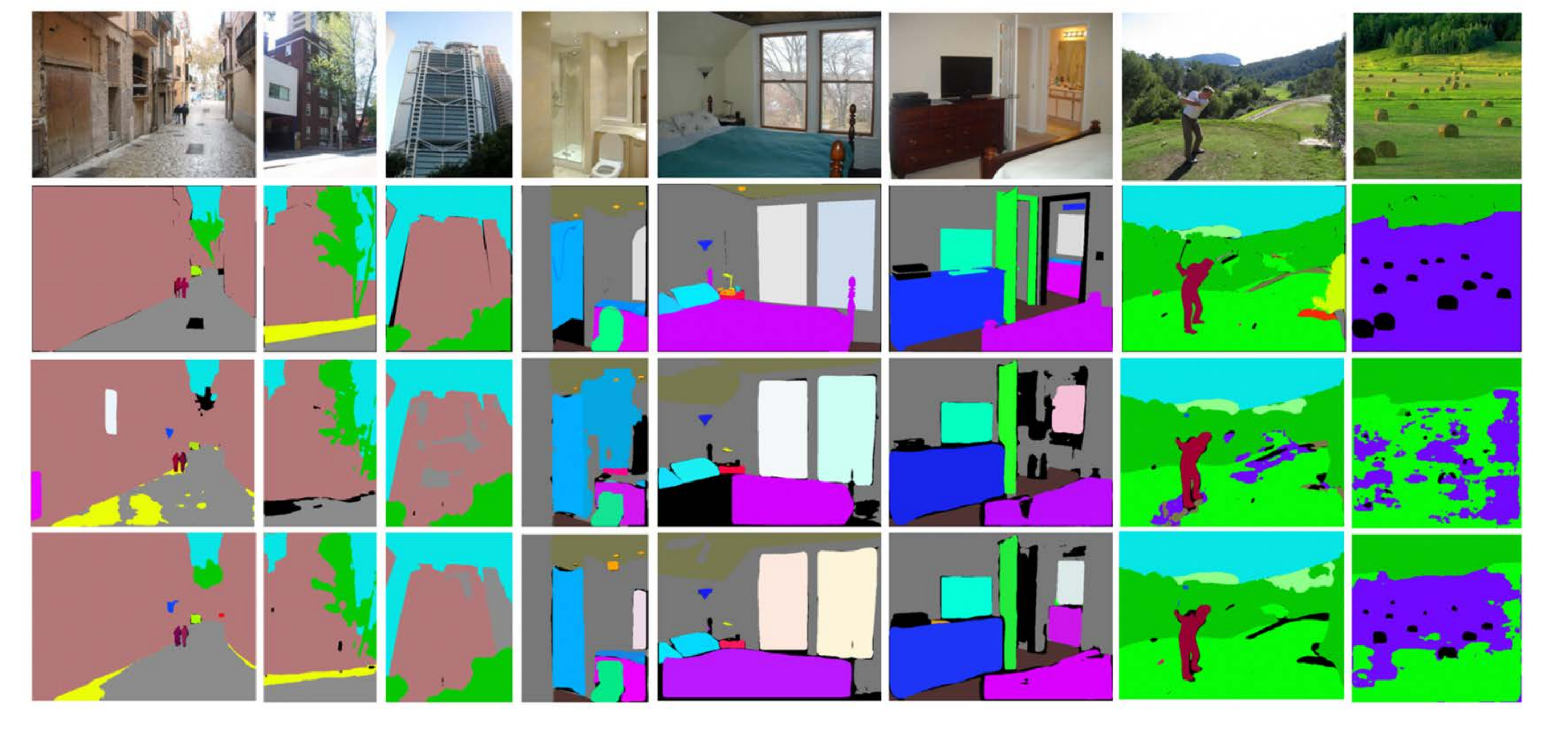}
    \caption{Visualized comparisons of panoptic segmentation outputs on ADE20K dataset. Raw images, Ground-Truth Segmentation, Panoptic-FPN outputs and BGRNet outputs are presented from top to bottom.}
    \label{fig:visualization}
\end{figure*}

The qualitative results on the ADE20K dataset are shown in Figure~\ref{fig:visualization}. As can be observed, our approach outputs more semantically meaningful and precise predictions than baseline methods despite the existence of complex object appearances and challenging background contents. For example, the baseline mistakes \textit{field} for \textit{grass} while our BGRNet predicts correctly thanks to the propagated information from the \textit{things} in the image. More visual results on COCO and ADE20K can be found in Supplementary Materials.

\begin{table}[ht]
\caption{Ablation studies on ADE20K val set.}
\label{tab:ablation_ade}
\centering
\begin{tabular}{l|lll}
\toprule[1pt]
Methods  & PQ & PQ$^{Th}$ & PQ$^{St}$  \\
\midrule[1pt]
Baseline      &30.1 &33.3    &23.7     \\   \hline
w Thing-Graph   & 30.6 &  33.7   & 24.9 \\         
w Stuff-Graph  &30.7    &33.0   &26.2          \\ 
w Thing-Graph/Stuff-Graph   &31.1 &33.5 &26.5  \\ \hline
Our BGRNet  &  31.8  & 34.1   &  27.3  \\
\bottomrule[1pt]
\end{tabular}
\end{table}

\subsection{Ablation Study}
\label{ablation}

\begin{table*}[ht]
\caption{Comparisons of different graphs and architectural designs on ADE20K val set. }
\label{tab:arch_design_ablation}
\centering
\scriptsize
\tabcolsep 0.02in 
\begin{tabular}{c|c|c|c|c|c|c|c|c|c|c|c|c}
\toprule[0.9pt]
\multirow{2}{*}{~~~\#~~~} & 
\multirow{2}{*}{Basic network~\cite{he2017mask}} & \multicolumn{2}{c|}{Thing-Graph Construction} & \multicolumn{2}{c|}{Stuff-Graph Construction} & \multicolumn{2}{c|}{Graph Connection} & \multicolumn{2}{c|}{Reasoning direction} & 
\multirow{2}{*}{PQ} & \multirow{2}{*}{PQ$^{Th}$} & \multirow{2}{*}{PQ$^{St}$} \\ \cline{3-10}
  & & Knowledge Graph \cite{jiang2018hybrid} & Attention & Non-local \cite{wang2018non} & Class-center & Semantic similarity & Attention & Thing-Stuff & Stuff-Thing & & & \\ \hline
  1 & \checkmark & & & & & & & & & 30.1     &33.3    &23.7 \\ \hline
  2 & \checkmark &\checkmark & & & & & & & & 30.4 & 33.5 & 24.2 \\ \hline
  3 & \checkmark & & \checkmark & & & & & & & 30.6 & 33.7 & 24.9 \\ \hline
  4 & \checkmark & & & \checkmark & & & & & &  30.6 & 32.8 & 26.3 \\ \hline
  5 & \checkmark & & & & \checkmark & & & & & 30.7 & 33.0 & 26.2 \\ \hline
  6 & \checkmark & & \checkmark & & \checkmark & \checkmark & & \checkmark & \checkmark & 31.5 & 33.7  & 27.1 \\ \hline
  7 & \checkmark & & \checkmark & & \checkmark & & \checkmark & \checkmark & & 31.4 & 33.6 & 27.0 \\ \hline
  8 & \checkmark & & \checkmark & & \checkmark & & \checkmark & & \checkmark & 31.6 & \textbf{34.3} & 26.2 \\ \hline
  9 & \checkmark & & \checkmark & & \checkmark & & \checkmark & \checkmark & \checkmark & \textbf{31.8} & 34.1 & \textbf{27.3} \\
\toprule[0.9pt]
\end{tabular}
\end{table*}

\begin{figure*}[ht]
    \centering
    \includegraphics[width=1.0\linewidth]{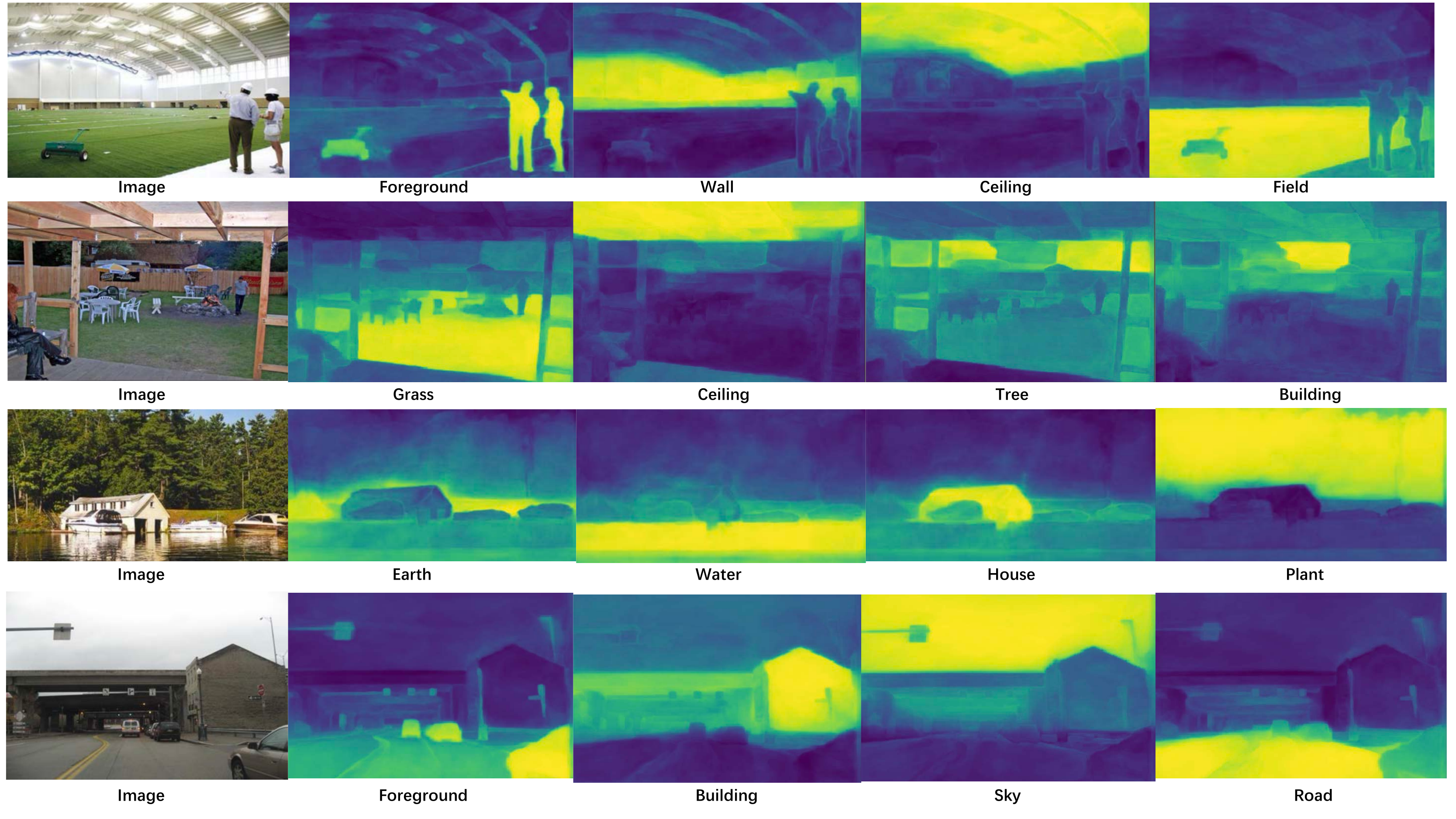}
    \caption{Visualization of similarities between extracted class centers and pixels generated by our method. Class Centers are listed below the images. The deeper the color is, the stronger the similarity between the class center and pixels. Benefited from the Class-center Stuff-Graph Construction scheme, our BGRNet can refine the local features under the guidance of the class center from a global view. Best viewed in color.}
    \label{fig:cls_center}
\end{figure*}

\noindent\textbf{Combinations of intra-modular and inter-modular graphs.}
Table ~\ref{tab:ablation_ade} shows the performance of different components of our BGRNet on ADE20K val set. 
``w Thing(Stuff)-Graph'' only has a single graph for foreground or background branch, while ``w Thing-Graph/Stuff-Graph'' contains graphs in both two branches with no inter-branch interaction, and the graph nodes are re-projected to visual features similar to Section \ref{sec_proj}.

We first analyze the effect of a single graph in either \textit{things} branch or \textit{stuff} branch. 
For single Thing-Graph, both PQ$^{Th}$ and PQ$^{St}$ get improved thanks to the region-wise reasoning that considers the correlations among proposals. 
For single Stuff-Graph, PQ$^{St}$ got a 2.5\% relative improvement, which showcases the great effect of extracting class centers to refine local features in a coarse-to-fine paradigm. 
Incorporating these two graphs with no connection across branches, the overall PQ is already 1\% higher than the baseline, which is a considerable improvement on challenging ADE20K dataset. 
Furthermore, we introduce graph connection module, which greatly improves the segmentation quality of \textit{things} and \textit{stuff}, due to the ability to mine the underlying relations between foreground and background.
As can be seen from the last row in Table \ref{tab:ablation_ade}, our BGRNet improves PQ$^{Th}$ and PQ$^{St}$ by 0.8\% and 3.6\% respectively, resulting in 31.8\% overall PQ, which outperforms Panoptic-FPN \cite{kirillov2019panopticFPN} by a large margin.

\noindent\textbf{Thing/Stuff-Graph Construction.}
To validate the efficiency of the proposed Thing-Graph and Stuff-Graph, we consider different construction methods and compare their performance in Table \ref{tab:arch_design_ablation}(\#2,\#3). 
Regarding Thing-Graph, we consider establishing the region-wise relations via a fixed knowledge graph. 
As for the knowledge graph for foreground objects, we follow \cite{jiang2018hybrid} to construct a fixed relation knowledge Thing-Graph and extract an adjacency matrix of regions according to their class predictions. 
This scheme achieves 30.4\% PQ, which is inferior to the adopted multi-head attention mechanism in BGRNet. 
The weakness may lie in the wrong region graphs due to the misclassification of some proposals, which indicates that the edge weights between some proposals are not reasonable anymore. 
As for the non-local graph for background, though slightly higher PQ$^{St}$ (26.3\% vs 26.2\%) is achieved, it incurs much larger computation since every pixel is regarded as a graph node.
Furthermore, with a non-local graph, the subsequent graph connection will be prohibitively expensive when the region-based Thing-Graph is considered.
As can be seen, constructions of attention-based Thing-Graph and class-center Stuff-Graph lead to higher performance and moderate computation.

\noindent\textbf{Different Graph Connection matrices.}
We also investigate the performance of our model using a different graph connection method, i.e., semantic similarity. 
To be exact, the $\tilde{\textbf{A}}$ in Equation \ref{eq_block} is built on the semantic similarity other than a multi-head mechanism under this setting.
The word embeddings of predicted classes of regions and stuff names of class centers are used to calculate the cosine similarity to form an adjacency matrix. 
As can be seen in Table~\ref{tab:arch_design_ablation}, the semantic similarity-based connection is also helpful in bridging the chasm between \textit{things} and \textit{stuff} and achieves 31.5\% PQ, which is still lower than that of attention-based mechanism (31.8\% PQ).
This indicates that our Graph Connection Module is supposed to obtain more sufficient expressive power and discover the diverse relations between \textit{things} nodes and \textit{stuff} nodes in a complicated scene than merely depends on a fixed linguistic graph.

\noindent\textbf{Unidirectional enhancement.}
We investigate the direction of Graph Connection by exploring unidirectional enhancement in Table~\ref{tab:arch_design_ablation}.
Previous method \cite{Li_2019_Aunet} uses two sources of attention to perform unidirectional enhancement from the foreground branch to background branch. 
To fully leverage the reciprocal relations between foreground and background, we thus investigate and compare the performance with different enhance directions. 
`Thing-Stuff' stands for only enhancing the feature of semantic segmentation branch after Graph Connection. `Stuff-Thing' represents only enhancing the feature of detection branch after Graph Connection.
It can be found that although unidirectional enhancement can lead to considerable performance gain, merely performing Graph Connection in one direction is not able to fully enhance the feature, and a two-way graph connection further boosts the overall PQ to 31.8\%.

\noindent\textbf{Visualize the correlations.}
To demonstrate the representative characteristics  of the extracted class centers described in Section \ref{sec_graph_repre}, we visualize the similarity between particular \textit{stuff} class centers and local features of pixels in Figure \ref{fig:cls_center}. 
As can be seen, the extracted \textit{stuff} class center correlates well with corresponding area and the responses in other area are inhibited, despite the existence of multiple \textit{stuff} classes, class ambiguity and fuzzy edges between different \textit{stuff} classes. 
For example, in the third row, 
the extracted class centers correlate well with the confusing \textit{stuff} class including \textit{plant}, \textit{water} and \textit{earth}.
Under the guidance of the class center features from a global view, local features can be refined.
This greatly improves the performance of our model in terms of PQ$^{St}$.

\section{Conclusion}
This paper introduces a Bidirectional Graph Reasoning Network (BGRNet) for panoptic segmentation that simultaneously segments foreground objects at the instance level and parses background contents at the class level. We propose a Bidirectional Graph Connection Module to propagate the information encoded from the semantic and co-occurrence relations between \textit{things} and \textit{stuff}, guided by the appearances of the objects and the extracted class centers in an image.  Extensive experiments demonstrate the superiority of our BGRNet, which achieves the new state-of-the-art performance on two large-scale benchmarks.

\section{Acknowledgement}
This work was supported in part by National Key R\&D Program of China under Grant No. 2018AAA0100300, National Natural Science Foundation of China (NSFC) under Grant No.U19A2073 and No.61976233, Nature Science Foundation of Shenzhen Under Grant No. 2019191361.

{\small
\bibliographystyle{ieee_fullname}
\bibliography{main}
}

\end{document}